\documentclass[10pt,twocolumn,letterpaper]{article}
\usepackage{cvpr}
\usepackage{multirow}
\definecolor{cvprblue}{rgb}{0.21,0.49,0.74}
\usepackage[pagebackref,breaklinks,colorlinks,allcolors=cvprblue]{hyperref}
\def\confName{CVPR}
\def\confYear{2027}

\makeatletter
\renewcommand{\paragraph}[1]{\par\vspace{0.45em}\noindent\textbf{#1}\hspace{0.8em}}
\makeatother

\title{STARS: Stage-Adaptive Responsibility Separation for Shadow Removal in Remote Sensing Images}

\author{\small Gaoliang Song$^{1\ast}$\quad Chengjie Li$^{1\ast}$\quad Pengju Liu$^{1}$\quad Chenchen Chang$^{2}$\quad Jinhao You$^{1}$\quad Hongzhi Zhang$^{1,\dagger}$\\[4pt]
\footnotesize $^{1}$ Harbin Institute of Technology\qquad $^{2}$ Northeast Forestry University\\[2pt]
\footnotesize $^{\ast}$ Co-first authors\qquad $^{\dagger}$ Corresponding author\\[2pt]
\footnotesize E-mail: 2023110047@stu.hit.edu.cn, genkliu007@gmail.com}

\begin{document}
\maketitle

\begin{abstract}
Shadows are a ubiquitous radiative attenuation phenomenon in remote sensing images and can severely affect high-resolution remote sensing interpretation tasks. In shadow removal, although introducing a mask prior provides explicit positional information for the restoration network, it also causes non-shadow regions to be modulated by the shadow prior, producing a systematic brightness shift; notably, this phenomenon is independent of whether the mask is accurate. In remote sensing images, non-shadow regions usually occupy a larger proportion, so a small shift over these regions incurs a much higher cost on whole-image metrics than an equal gain in shadow regions. Adopting a task-decoupling perspective, we therefore decompose shadow removal into two tasks, shadow-region restoration and non-shadow-region identity, within a cascaded architecture, and handle them sequentially. Based on this insight, we propose STARS, a cascaded shadow removal framework that separates regional responsibilities across stages. The method retains the advantages of the shadow mask while resolving the degradation of non-shadow regions, without requiring a complex parallel dual-branch structure. Meanwhile, to introduce large-model prior knowledge into remote sensing tasks, we further design a Feature Selection and Fusion Adapter (FSFA). We conduct extensive experiments on three benchmark datasets, UAV-SC+, ISTD+ and WSRD+ 2026, demonstrating the superiority of STARS. The framework shows that decoupling regional responsibilities can resolve the intrinsic contradiction of mask priors, providing a new direction for image restoration with region heterogeneity.
\end{abstract}

\section{Introduction}

Shadows are a radiative attenuation phenomenon caused by objects occluding light \cite{ref1,ref2}. In high-resolution optical remote sensing images, shadows weaken the radiance information within shadowed areas, compress the dynamic range and alter the spectral distribution, posing significant interference to downstream tasks such as object detection and semantic segmentation \cite{ref3,ref1,ref2,ref4}. Shadow removal is therefore an important research direction in remote sensing image processing \cite{ref1,ref2}. In recent years, deep-learning-based shadow removal has advanced rapidly \cite{ref1,ref2,ref5}. The shadow mask provides the most direct positional prior for the removal network: the detected shadow region exactly indicates the area to be restored, and introducing it as an input channel or an attention condition can substantially improve restoration quality in shadowed regions \cite{ref6,ref7,ref8}.

\begin{figure}[t]
  \centering
  \includegraphics[width=\linewidth]{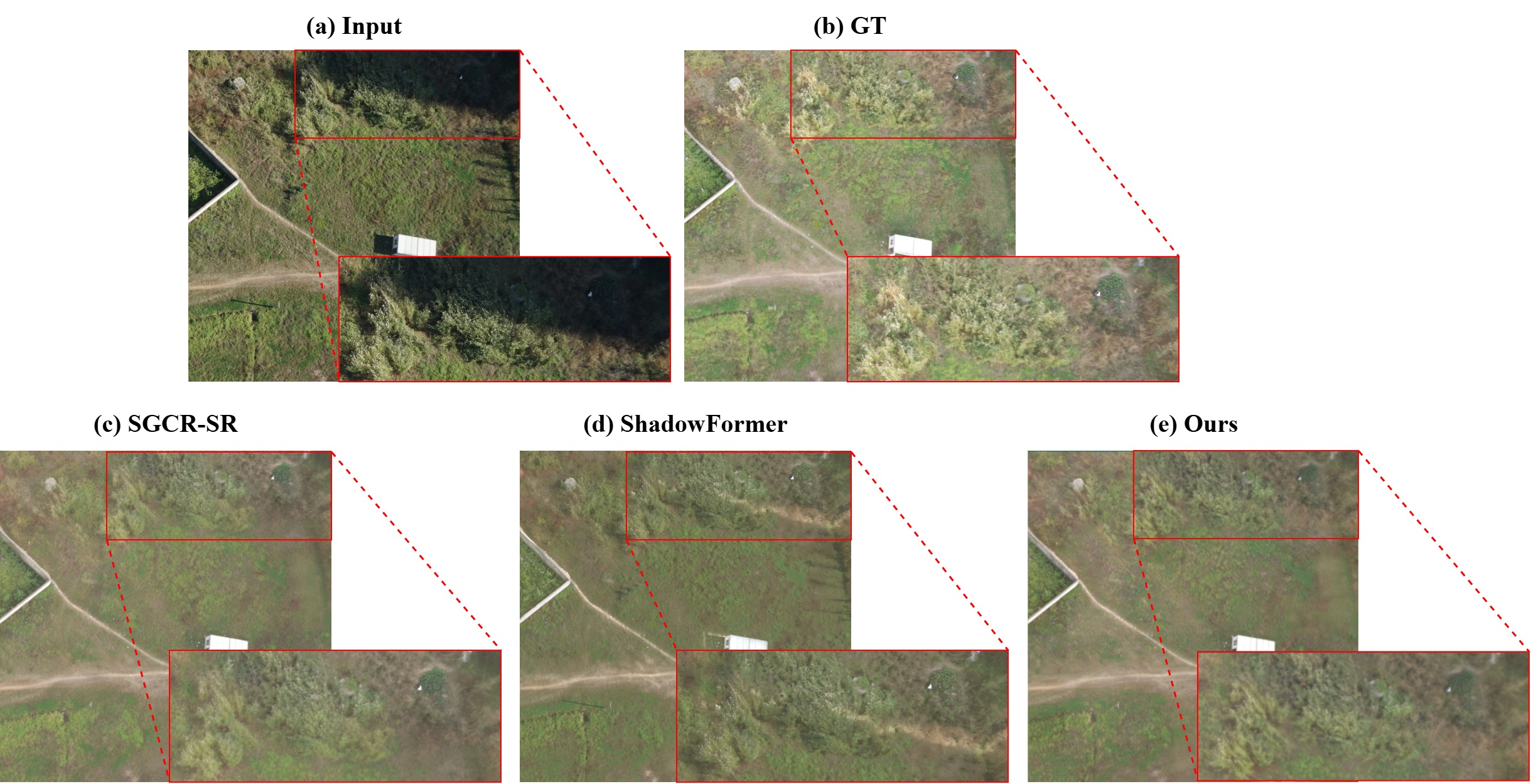}
  \caption{Comparison between a method without a mask (SGCR-SR) and a method with a mask prior (ShadowFormer). ShadowFormer restores the shadow region better but exhibits an obvious brightness shift in the non-shadow region, whereas our method (Ours) achieves better restoration quality in both regions.}
  \label{fig1}
\end{figure}

However, models that use a mask prior often exhibit the following phenomenon: for mask-conditioned networks such as ShadowFormer, the shadow and non-shadow regions are optimized jointly by a single set of weights, the two tasks compete during optimization, and the fidelity of non-shadow regions degrades markedly; for models such as SGCR-SR that originally contain no mask prior, directly introducing the mask improves restoration in shadowed regions but induces a systematic brightness shift in non-shadow regions, as shown in Fig.~\ref{fig1}.

In remote sensing images, shadowed areas typically occupy only a small fraction of the whole image; sacrificing large non-shadow regions for better shadow restoration incurs losses far exceeding the gains. Methods that abandon the mask (e.g., OmniSR \cite{ref11}, SGCR-SR \cite{ref9}) circumvent this problem but also discard the positional prior, limiting shadow restoration. This paper therefore seeks to retain the spatial prior of the mask while systematically resolving its side effect on non-shadow regions. The key to this lies in decoupling shadow removal by region: the task of the shadow region is to remove the shadow, whereas the task of the non-shadow region is to remain unchanged, and the two conflicting regional tasks should be explicitly separated and accomplished sequentially.

We realize this decoupling with a cascaded architecture and the shadow mask. In a cascade, the output of the previous stage is the input of the next, so a complex task can naturally be split across stages, and each stage has independent parameters, so adjusting the responsibility of one stage does not alter the structure of the others. Meanwhile, the shadow mask is an explicit regional marker. Combining the two, we let the shadow mask serve as a regional task marker so that the two conflicting regional tasks are assigned to different stages of the cascade, realizing an explicit division of regional tasks. Based on this understanding, we propose a responsibility separation strategy, STARS (Stage-Adaptive Responsibility Separation), with the following contributions:

\begin{enumerate}[itemsep=1pt,parsep=0pt,topsep=2pt,partopsep=0pt]
\item We propose a cascaded responsibility separation strategy. This design turns the mask from a conventional spatial conditioning input into an explicit mechanism for coordinating regional restoration tasks across stages; the shadow removal task is decoupled into two regional sub-tasks that are assigned to different stages and accomplished sequentially.
\item We design a feature selection and fusion adapter that transfers DINOv3 semantic features, developed for natural images, to remote sensing shadow removal, offering a new solution for feature transfer and fusion between natural and remote sensing images.
\item We design a region-specific loss $\mathcal{L}_\mathrm{ns}$ for non-shadow regions, which guarantees at the optimization level that the assigned responsibilities are actually executed, complementing the structural division.
\end{enumerate}

\begin{figure*}[t]
  \centering
  \includegraphics[width=0.90\linewidth]{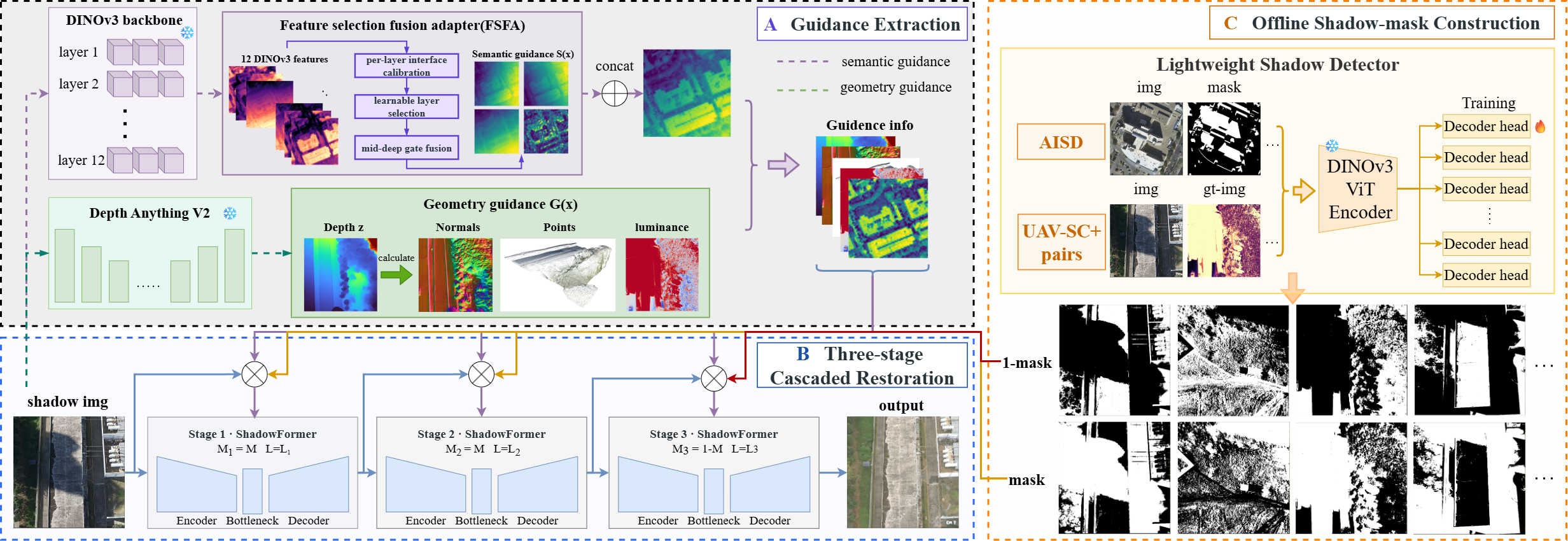}
  \caption{Overall framework. Our method consists of three parts: A, guidance extraction, which extracts semantic and geometric information with a frozen DINOv3 backbone and Depth Anything V2, respectively; C, an offline shadow mask detection head, a lightweight shadow detector trained on the AISD dataset and UAV-SC+ paired data, which extracts the shadow mask; B, a three-stage cascade composed of three consecutively connected ShadowFormers. STARS is embodied by the mask direction at each stage ($M$ / $1-M$) and the residual modulation direction, and $\mathcal{L}_\mathrm{ns}$ constrains only the non-shadow region of Stage 3.}
  \label{fig2}
\end{figure*}

\section{Related Work}

\subsection{Shadow Processing in Remote Sensing Images}

Shadow processing in remote sensing has evolved from physical models to data-driven methods. Early methods predict shadow extents using solar geometry parameters and digital elevation models (DEM) \cite{ref12,ref13}, or estimate illumination compensation in shadowed regions with complex irradiance models \cite{ref14,ref15}; feature-based methods extract shadow attributes in specific color spaces and texture domains and apply thresholds \cite{ref16}. These methods rely on high-accuracy auxiliary data or hand-crafted features, adapt poorly to complex urban environments, and often produce incomplete detections or hard reconstruction artifacts \cite{ref12,ref13,ref15}.

In recent years, deep learning has become the mainstream. Shadow detection in remote sensing follows binary semantic segmentation, and dense prediction frameworks such as U-Net with spatial attention have significantly improved detection accuracy in complex scenes \cite{ref5}. Remote sensing shadow removal advances along two lines: end-to-end networks without shadow detection, e.g., ESCNet \cite{ref1} builds the UAV-SC benchmark and performs restoration and radiometric correction with two subnets, and methods targeting shadow distribution properties, e.g., RMMamba \cite{ref2} balances shadow pixels via random rearrangement and a color prior, while RSMamba \cite{ref17} combines a Retinex prior with state-space modeling.

Methods in the natural image domain fall into two categories: mask conditioning (ST-CGAN \cite{ref6} takes the mask as input; ShadowFormer \cite{ref8} uses mask-weighted residuals and attention) and regional modeling (DMTN \cite{ref18} decouples detection and removal; RASM \cite{ref19} applies regional attention). Both complete the restoration of the two regions with shared weights; the mask prior degrades non-shadow fidelity, and the regional conflict remains unresolved at the structural level.
\subsection{Semantic and Geometric Guidance and Cascaded Refinement}

Task-agnostic representations from self-supervised visual foundation models are widely used as frozen guidance for restoration: DINOv2 \cite{ref20} demonstrates the robustness and transferability of self-supervised ViT features, DINOv3 \cite{ref21} extends the training to large-scale data, and Depth Anything v2 \cite{ref22} provides monocular depth from which point maps and normals can be derived; OmniSR \cite{ref11} and SGCR-SR \cite{ref9} introduce such cues into shadow removal as representative baselines. Progressive refinement, where later stages correct residual artifacts of earlier ones \cite{ref23,ref24}, is a classical restoration paradigm: ESCNet \cite{ref1} divides restoration and radiometric compensation into two stages, while Li et al. \cite{ref10} prove that shadow restoration and non-shadow identity mapping are incompatible under shared weights and propose parallel dual branches (IMB/IDB) to separate them.

\section{Method}

\subsection{Overall Framework}

\begin{figure}[t]
  \centering
  \includegraphics[width=0.88\linewidth]{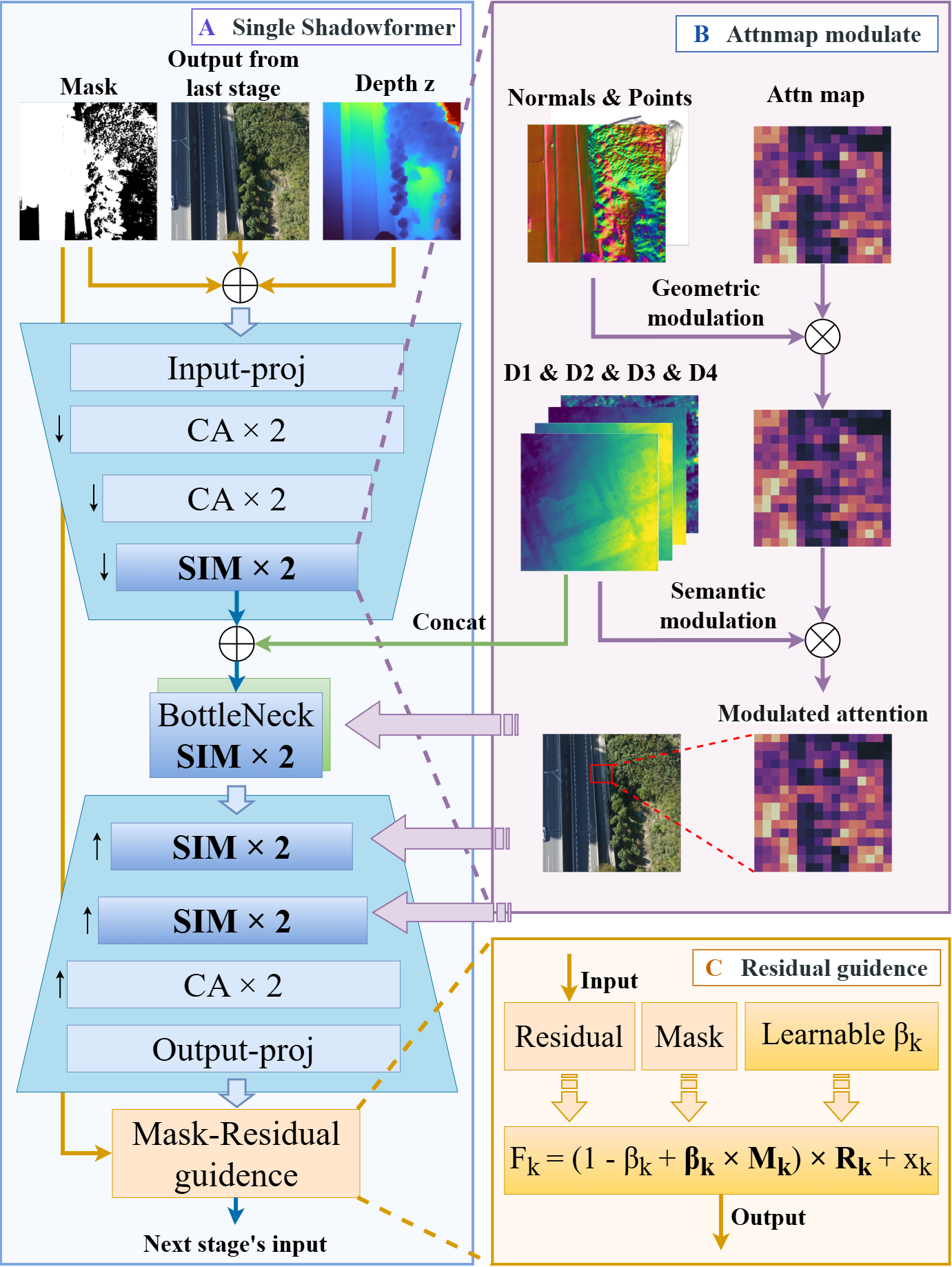}
  \caption{Structure of the improved ShadowFormer stage. (A) Single-stage backbone: the shadow mask serves as an additional input channel and participates in feature extraction together with the shadow image and depth $z$; (B) attention modulation: geometric cues and multi-scale semantic features modulate the SIM block attention, and semantic features are also concatenated into the bottleneck; (C) residual guidance: the backbone residual $R_k$ is weighted by the mask $M_k$ and a learnable coefficient $\beta_k$, realizing structural-level regional responsibility division.}
  \label{fig3}
\end{figure}

Given a shadow image $x \in \mathbb{R}^{H \times W \times 3}$, the goal is to reconstruct a shadow-free image $\hat{y} \approx y^{*}$. We use a three-stage refinement cascade in which each stage keeps the ShadowFormer \cite{ref8,ref9} backbone; a single stage is shown in Fig.~\ref{fig3}.
Let $S(x)$ denote the semantic features extracted from the input; $G(x) = \{z,\, P(x),\, N(x)\}$ the geometric guidance, where $z$ is the normalized depth channel concatenated with RGB, and $P(x)$ and $N(x)$ are the point map and surface normal obtained by back-projecting depth, respectively; $M$ denotes the soft shadow mask. Let $f_k(\cdot)$ denote the $k$-th stage backbone; the cascade is defined as
\begin{equation}
\begin{aligned}
  \hat{y}^{(1)} &= f_1\big(x,\, S(x),\, G(x),\, M\big), \\
  \hat{y}^{(k)} &= f_k\big(\hat{y}^{(k-1)},\, S(x),\, G(x),\, M_k\big),\quad k = 2, 3,
\end{aligned}
  \label{eq:cascade}
\end{equation}
with the final prediction $\hat{y} = \hat{y}^{(3)}$. Unlike previous work that uses the mask merely as a positional prior, our model lets the mask carry cross-stage responsibility assignment: stages receive masks in different forms ($M$ or $1-M$), and the output residuals are gated by the mask in different directions, turning the stage division from an implicit, data-driven formation into an explicit structural specification. The overall framework is shown in Fig.~\ref{fig2}.

\subsection{STARS: Cascaded Responsibility Separation Strategy}

As discussed in the introduction, injecting the mask into a shared-weight network causes a systematic brightness shift in non-shadow restoration that cannot be effectively alleviated. We therefore change the strategy: instead of trying to prevent the brightness shift in non-shadow regions, we allow it to appear first and then decouple the correction of the brightness shift from shadow removal as a new task assigned to the last stage. In the three-stage cascade used here, this decoupling and division are guided by a bidirectional shadow mask.
\subsubsection{Offline Acquisition of the Shadow Mask}
\label{sssec:mask}

The core of cascaded responsibility separation is the shadow mask $M$. Since remote sensing scenes lack manual shadow annotations and no ground-truth mask exists at inference time, $M$ can only be estimated from the shadow image itself, independently of the removal network, so as to avoid coupling with model parameters. We generate shadow probability maps with a frozen DINOv3 encoder and a lightweight decoding head: the head is first pretrained on the annotated AISD dataset \cite{ref25}, and then domain-adapted via proxy supervision constructed from the paired remote sensing dataset UAV-SC+ \cite{ref1}. At inference, only the shadow image is fed, and the output probability map is binarized and softened to obtain $M$. The mask is generated offline and kept fixed as a conditional input to the removal network; it does not participate in its training or inference updates.
\subsubsection{Bidirectional Mask Input Channels}

Introducing the shadow mask as an input channel is the lightest way to preserve the positional prior: at every stage the network directly sees where to operate, without extra branches or structural changes.

The soft mask $M\in[0,1]$ obtained in Sec.~\ref{sssec:mask} enters the main restorer as the regional condition. The regional conditions of the three stages are defined as
\begin{equation}
  M_1 = M_2 = M, \qquad M_3 = 1 - M .
  \label{eq:region}
\end{equation}
Hence stages 1--2 take $[F^{(k-1)},\, z,\, M]$ as input, and stage 3 takes $[F^{(2)},\, z,\, 1-M]$. The mask is concatenated after depth $z$; the first four RGB-D channels remain unchanged, and the input-layer weights corresponding to the added channel are zero-initialized.

This design gives the mask different semantics across stages: for the first two stages, $M=1$ marks the shadow region to be restored, so restoration focuses there; for the last stage, $1-M=0$ hides the already-restored shadow region from its input, while marking the non-shadow region to be corrected. The mask thus serves as both a positional prior and a cross-stage task-allocation signal, separating regional responsibilities structurally.
\subsubsection{Bidirectional Mask Residual Guidance}

Input channels alone are not sufficient to realize the division of labor at each stage. We further apply a learnable mask gate to the residual at the output of each stage:
\begin{equation}
  F_k = \big( 1 - \beta_k + \beta_k \cdot M_k \big) \odot R_k + x_k ,
  \label{eq:residual}
\end{equation}
where $R_k$ is the residual predicted by the $k$-th backbone, $x_k$ is the stage input, and $\beta_k$ is a learnable scalar per stage.
The gating direction follows the regional responsibility of the stage: the first two stages use the shadow-region residual with a ratio tending to a constant, concentrating restoration energy inside shadows; the last stage fully uses the non-shadow residual and excludes the shadow residual, which cannot interfere with the regions already restored. Compared with feature-level modulation, this position produces the missing restoration information rather than scaling corrupted features. At initialization the gate equals the original residual connection, preserving the backbone's refinement capability, and $\beta$ has non-zero gradients at the initial point, so each stage can learn its guidance strength.
\begin{figure*}[t]
  \centering
  \includegraphics[width=0.90\linewidth]{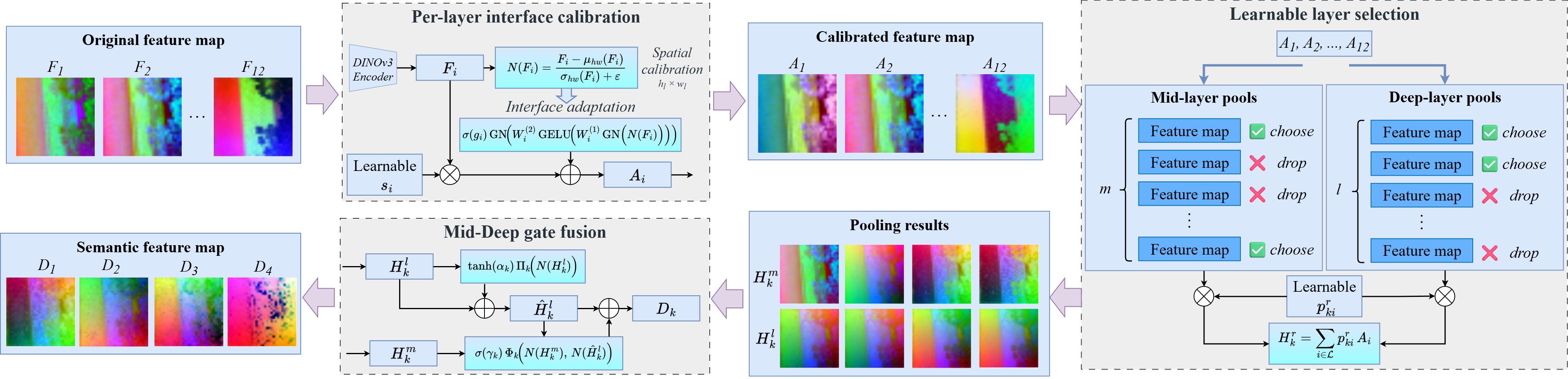}
  \caption{Structure of the semantic guidance adapter. The adapter first aligns the distributions of the features of each DINOv3 layer, then softly selects feature layers from the mid-level and deep pools with learnable weights, and finally fuses mid-level details with deep semantics through gated residuals.}
  \label{fig4}
\end{figure*}

\subsubsection{Non-Shadow Region-Specific Supervision}

The structural division specifies the last stage's responsibility architecturally, but the non-shadow shift is small: with the global loss alone, the gradient vanishes as the error approaches zero and the last stage tends to make no correction. We therefore introduce non-shadow-region-specific supervision for the last stage:
\begin{equation}
  \mathcal{L}_{\mathrm{ns}} = \big\| (1 - M) \odot (F_3 - y^{*}) \big\|_1 ,
  \label{eq:nsloss}
\end{equation}
This loss counts only non-shadow regions and excludes the shadow region from the gradient, so the last stage's signal comes entirely from its responsibility scope without disturbing the first two stages; the L1 form keeps the gradient constant near zero and provides a strong drive for non-shadow regions.
\subsection{Feature Selection and Fusion Adapter}

Shadow contexts in remote sensing images are diverse, and the representation of the restoration network alone can hardly provide sufficient semantic cues. Large models pretrained on natural images offer strong semantic priors, but this prior cannot be introduced directly into the cascade: existing restoration networks were trained on the feature distributions of specific semantic backbones, and the output features of different backbones are biased; direct connection would substantially degrade performance \cite{ref20,ref21}. To introduce the frozen DINOv3 semantic prior into remote sensing shadow removal, we design a feature selection and fusion adapter for the intermediate layers of ShadowFormer, whose structure is shown in Fig.~\ref{fig4}.

The adapter first calibrates the features $F_i$ of each DINOv3 layer with per-image statistics (Eq.~\eqref{eq:norm}) to capture cross-layer distribution differences, then performs interface adaptation with a low-gain bottleneck residual (Eq.~\eqref{eq:interface}), obtaining calibrated features $A_i$ and separating shallow and deep semantics.
\begin{equation}
  N(F_i) = \frac{F_i - \mu_{hw}(F_i)}{\sigma_{hw}(F_i) + \varepsilon},
  \label{eq:norm}
\end{equation}
where $\mu_{hw}(F_i)$ and $\sigma_{hw}(F_i)$ are the mean and standard deviation computed over the $h_i\times w_i$ spatial positions of \textbf{the same image and channel}, respectively; $\varepsilon$ is a positive stability term.
\begin{equation}
  \resizebox{0.9\columnwidth}{!}{$\displaystyle A_i = s_i F_i + \sigma(g_i)\, \mathrm{GN}\!\big( W_i^{(2)} \, \mathrm{GELU}\!\big( W_i^{(1)} \, \mathrm{GN}\big( N(F_i) \big) \big) \big)$}
  \label{eq:interface}
\end{equation}
where $s_i$ is a layer-wise learnable scaling coefficient; $W_i^{(1)}$ and $W_i^{(2)}$ are the dimension-reduction and dimension-expansion projections, respectively; $\mathrm{GN}(\cdot)$ and $\mathrm{GELU}(\cdot)$ denote group normalization and nonlinear reparameterization; $g_i$ is the learnable gate bias of the bottleneck residual branch; and $\sigma(\cdot)$ is the sigmoid function.
This transformation aligns the translation and scale of each layer's features, so that layers can be fused in the shared interface while preserving the spatial distribution within each channel.

DINOv3 has many intermediate layers with different semantics; rather than hard-specifying several layers, we set learnable pooling weights $p_{ki}^{r}$ with $p_{ki}^{r}\geq0$ and $\sum_{l\in\mathcal{L}}p_{ki}^{r}=1$ (Eq.~\eqref{eq:pool}), so that the adapter learns which layers to select and obtains $H_k^r$.
\begin{equation}
\begin{aligned}
  H_k^{r} &= \sum_{i \in \mathcal{L}} p_{ki}^{r}\, A_i, \\
  p_{ki}^{r} &= \frac{\exp\!\big( a_{ki}^{r} / T \big)}{\sum_{i \in \mathcal{I}} \exp\!\big( a_{ki}^{r} / T \big)},\quad r \in \{m, l\},
\end{aligned}
  \label{eq:pool}
\end{equation}
where $k$ denotes the $k$-th adapter output; $H_k^m$ and $H_k^l$ denote the pooled mid-level detail and deep-level semantic results of the $k$-th output, respectively; and $a_{ki}^{r}$ is a learnable logit for output $k$, candidate layer $i$ and pooling type $r$.

Selected layers spread over the deep pool $l$ and the mid-level pool $m$: deep features are semantically strong but spatially coarse, whereas mid-level features are detail-rich but less abstract. A gated residual $\sigma(\gamma_k)$ injects mid-level details into the deep semantic map so that the two levels complement each other (Eq.~\eqref{eq:mix}), and the resulting $D_k$ modulates attention.
\begin{equation}
\begin{aligned}
  \hat{H}_k^{l} &= H_k^{l} + \tanh(\alpha_k)\, \Pi_k\!\big( N(H_k^{l}) \big), \\
  D_k &= \hat{H}_k^{l} + \sigma(\gamma_k)\, \Phi_k\!\big( N(H_k^{m}),\, N(\hat{H}_k^{l}) \big),
\end{aligned}
  \label{eq:mix}
\end{equation}
where $\Pi_k$ is a projection residual branch acting only on $H_k^l$, and $\Phi_k$ is the mid-deep fusion branch.

\section{Experiments}
\label{sec:exp}

\subsection{Experimental Settings}
\label{sec:exp:setup}

\paragraph{Datasets} We conduct experiments on three challenging datasets, UAV-SC+ \cite{ref1}, ISTD+ \cite{ref6,ref26} and WSRD+ 2026 \cite{ref27,ref28}, to validate the effectiveness of our method.
\paragraph{UAV-SC+} UAV-SC+ is a UAV remote sensing shadow correction benchmark refined from UAV-SC \cite{ref1}. It provides paired shadow images and corresponding shadow-free reference images, with $5{,}771$ pairs for training and $30$ pairs for testing.
\paragraph{ISTD+} ISTD+ is a classic benchmark for natural-scene shadow removal, obtained by color-correcting ISTD \cite{ref6,ref26}. It consists of shadow images, shadow masks and shadow-free reference images, with $1{,}330$ training groups and $540$ testing groups.
\paragraph{WSRD+ 2026} WSRD+ 2026 is the benchmark adopted by the CVPR 2026 NTIRE Image Shadow Removal Challenge \cite{ref27}, evolved from the WSRD series of datasets \cite{ref28}. It contains $836$ and $150$ pairs for training and testing, respectively.
\paragraph{Evaluation Metrics} Following prior work \cite{ref8,ref9,ref10}, we report three widely used objective metrics: PSNR, SSIM \cite{ref29} and LPIPS \cite{ref30}. PSNR measures pixel-level fidelity, SSIM measures structural similarity in terms of luminance, contrast and structure, and LPIPS reflects texture details and perceptual naturalness.

Moreover, the core of our method is to resolve the side effect of the mask prior on non-shadow regions, and reporting whole-image metrics alone would conceal the performance difference between shadow and non-shadow regions. To reflect the behavior of the proposed method under practical deployment conditions, we further report PSNR on the shadow and non-shadow regions separately according to the shadow mask. In the following tables, S., N.S. and All denote the metrics on the shadow region, the non-shadow region and the whole image, respectively.
\begin{table*}[t]
  \centering
  \small
  \caption{Quantitative comparison with representative methods on UAV-SC+ and ISTD+. PSNR is reported on the shadow region (S.), the non-shadow region (N.S.) and the whole image (All) at $256\times256$; SSIM and LPIPS are computed on the whole image. Best results are in bold.}
  \label{tab:uavsc}
  \begin{tabular}{@{}p{0.49\textwidth}@{\hspace{2pt}}|@{\hspace{2pt}}p{0.49\textwidth}@{}}
    \centering
    \begin{minipage}[t]{0.49\textwidth}\centering
      \textbf{UAV-SC+ dataset}\\[2pt]
      {\footnotesize\renewcommand{\arraystretch}{0.84}%
      \setlength{\tabcolsep}{1.4pt}%
      \begin{tabular*}{\textwidth}{@{\extracolsep{\fill}}lccccc@{}}
      \toprule
      \multirow{2}{*}{Method} & \multicolumn{3}{c}{PSNR (dB) $\uparrow$} & \multirow{2}{*}{SSIM $\uparrow$} & \multirow{2}{*}{LPIPS $\downarrow$} \\
      \cmidrule(lr){2-4}
       & S. & N.S. & All & & \\
      \midrule
Mask-ShadowGAN \cite{ref31}& 27.57 & 20.45 & 19.13 & 0.737 & 0.5754 \\
    G2R-ShadowNet \cite{ref32}  & 24.69 & 17.98 & 16.83 & 0.718 & 0.3376 \\
    DMTN \cite{ref18}           & 31.60 & 24.15 & 23.02 & 0.819 & 0.3059 \\
    DHAN \cite{ref33}           & 33.06 & 25.17 & 24.13 & 0.862 & 0.3005 \\
    ShadowFormer \cite{ref8}    & 32.71 & 24.37 & 23.44 & 0.835 & 0.3053 \\
    TBRNet \cite{ref34}         & 31.22 & 23.66 & 22.57 & 0.772 & 0.3103 \\
    NeFour \cite{ref35}         & 33.11 & 25.32 & 24.25 & 0.872 & 0.2989 \\
    RASM \cite{ref19}           & 33.34 & 25.73 & 24.63 & 0.874 & 0.2994 \\
    RSMamba \cite{ref17}        & 33.20 & 25.48 & 24.41 & 0.874 & 0.2774 \\
    \midrule
    Ours                       & \textbf{34.57} & \textbf{28.08} & \textbf{26.65} & \textbf{0.897} & \textbf{0.0772} \\
      \bottomrule
      \end{tabular*}}
    \end{minipage}
    &
    \begin{minipage}[t]{0.49\textwidth}\centering
      \textbf{ISTD+ dataset}\\[2pt]
      {\footnotesize\renewcommand{\arraystretch}{0.93}%
      \setlength{\tabcolsep}{1.4pt}%
      \begin{tabular*}{\textwidth}{@{\extracolsep{\fill}}lccccc@{}}
      \toprule
      \multirow{2}{*}{Method} & \multicolumn{3}{c}{PSNR (dB) $\uparrow$} & \multirow{2}{*}{SSIM $\uparrow$} & \multirow{2}{*}{LPIPS $\downarrow$} \\
      \cmidrule(lr){2-4}
       & S. & N.S. & All & & \\
      \midrule
SP+M-Net \cite{ref26}       & 36.16 & 35.84 & 32.33 & 0.9479 & 0.0716 \\
    Param+M+D-Net \cite{ref36}  & 33.59 & 34.33 & 30.12 & 0.9420 & 0.0759 \\
    Fu et al. \cite{ref37}      & 36.41 & 31.01 & 29.43 & 0.8404 & 0.1673 \\
    LG-ShadowNet \cite{ref38}   & 32.65 & 33.36 & 29.20 & 0.9335 & 0.0920 \\
    DC-ShadowNet \cite{ref39}   & 32.20 & 33.21 & 28.76 & 0.9219 & 0.1112 \\
    G2R-ShadowNet \cite{ref32}  & 34.01 & 34.62 & 30.49 & 0.9330 & 0.0868 \\
    BMNet \cite{ref40}          & 37.30 & 35.06 & 32.30 & 0.9551 & 0.0567 \\
    SG-ShadowNet \cite{ref41}   & 37.41 & 34.95 & 32.41 & 0.9524 & 0.0594 \\
    \midrule
    Ours                        & \textbf{38.72} & \textbf{37.47} & \textbf{34.22} & \textbf{0.9612} & \textbf{0.0206} \\
      \bottomrule
      \end{tabular*}}
    \end{minipage}
  \end{tabular}
\end{table*}

\subsection{Implementation Details}
\label{sec:exp:impl}

\begin{figure}[t]
  \centering
  \includegraphics[width=\linewidth]{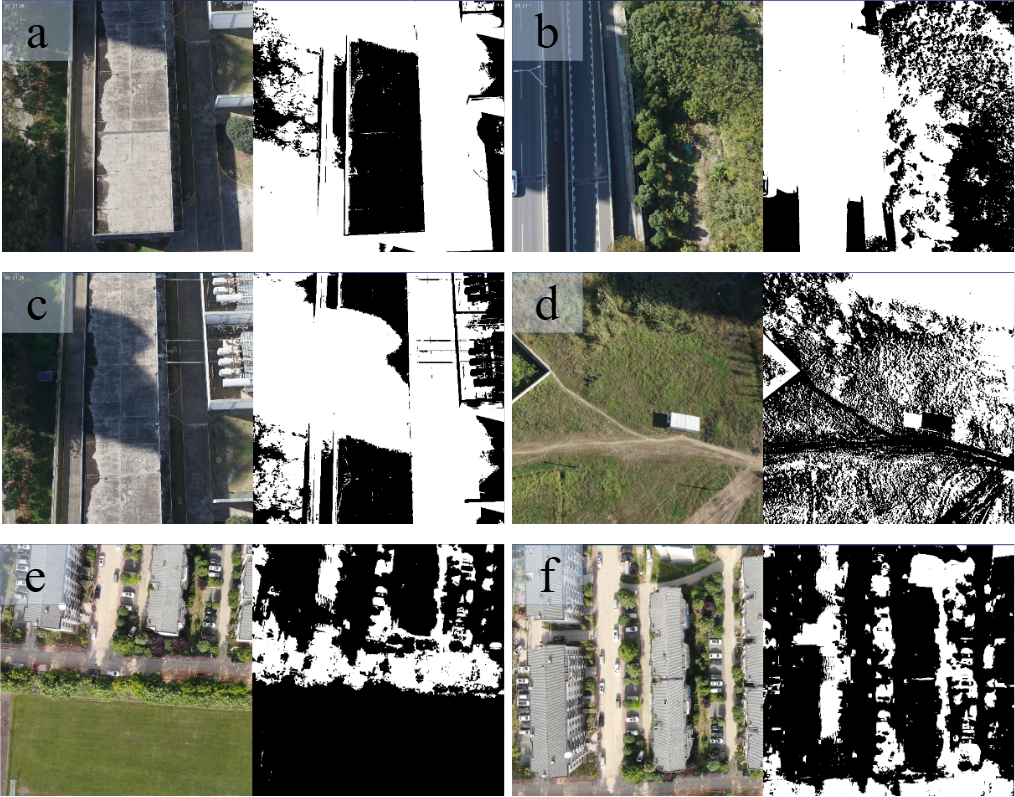}
  \caption{Examples of shadow mask estimation by the shadow detection head on UAV-SC+ images. After fine-tuning on UAV-SC+ shadow/non-shadow image pairs, the head achieves IoU 0.8607, F1 0.9252 and recall 0.9477 on the AISD dataset, indicating that it outputs trustworthy shadow masks.}
  \label{fig:mask_example}
\end{figure}

\paragraph{Shadow Mask Extraction} On the AISD dataset, we resize images to $512\times512$ and train a lightweight shadow detection head on the intermediate features of layers 4, 8, 12, 16, 20 and 23 of a frozen DINOv3 \cite{ref21} ViT-L/16 backbone for 100 epochs, with batch size 4, initial learning rate $2\times10^{-4}$, weight decay $1\times10^{-4}$, and cosine annealing to 5\% of the initial learning rate. To better adapt the head to remote sensing images, we select 2{,}157 shadow/shadow-free image pairs from UAV-SC+ \cite{ref1}, obtain shadow masks by image differencing, and fine-tune the head. During training and inference on UAV-SC+, the output of the head is binarized and softened, and then used as the conditional input to the main restorer and for region-wise evaluation; an example of mask estimation is shown in Fig.~\ref{fig:mask_example}.

\paragraph{Cascade Training} All experiments are conducted on a single NVIDIA RTX A6000 GPU. We adopt three-stage progressive training with the AdamW optimizer: on UAV-SC+, the first, second and third stages are trained for 150, 250 and 90 epochs, which takes about 10 days in total, with learning rates of $1\times10^{-4}$, $1\times10^{-4}$ and $5\times10^{-5}$, which are cosine-annealed to $5\times10^{-5}$, $5\times10^{-5}$ and $1\times10^{-6}$, respectively, and the responsibility separation strategy takes full effect only from the third stage. For ISTD+ and WSRD+ 2026, the numbers of training epochs are scaled according to dataset size. The loss function adopts different configurations at different stages, as shown in Eq.~\eqref{eq:total}. Since UAV-SC+ and WSRD+ 2026 have no officially annotated shadow masks, we use the output of the shadow detection head; for ISTD+, the official shadow masks are used. The training objectives of the stages are defined as
\begin{equation}
\begin{aligned}
  \mathcal{L}_1 &= \mathcal{L}_{\mathrm{mse}} + \lambda_{\mathrm{lpips}} \mathcal{L}_{\mathrm{lpips}}, \\
  \mathcal{L}_2 &= \mathcal{L}_1 + \lambda_{\mathrm{hessian}} \mathcal{L}_{\mathrm{hessian}} + \lambda_{\mathrm{stage}} \mathcal{L}_{\mathrm{stage}} \\
  &\quad + \lambda_{\mathrm{contraction}} \mathcal{L}_{\mathrm{contraction}}, \\
  \mathcal{L}_3 &= \mathcal{L}_2 + \lambda_{\mathrm{ns}} \mathcal{L}_{\mathrm{ns}},
\end{aligned}
  \label{eq:total}
\end{equation}
where $\mathcal{L}_1$, $\mathcal{L}_2$ and $\mathcal{L}_3$ are the training losses of phase 1, phase 2 and phase 3, respectively.

\subsection{Comparison with State-of-the-Art Methods}
\label{sec:exp:sota}

\begin{figure*}[t]
  \centering
  \includegraphics[width=0.90\linewidth]{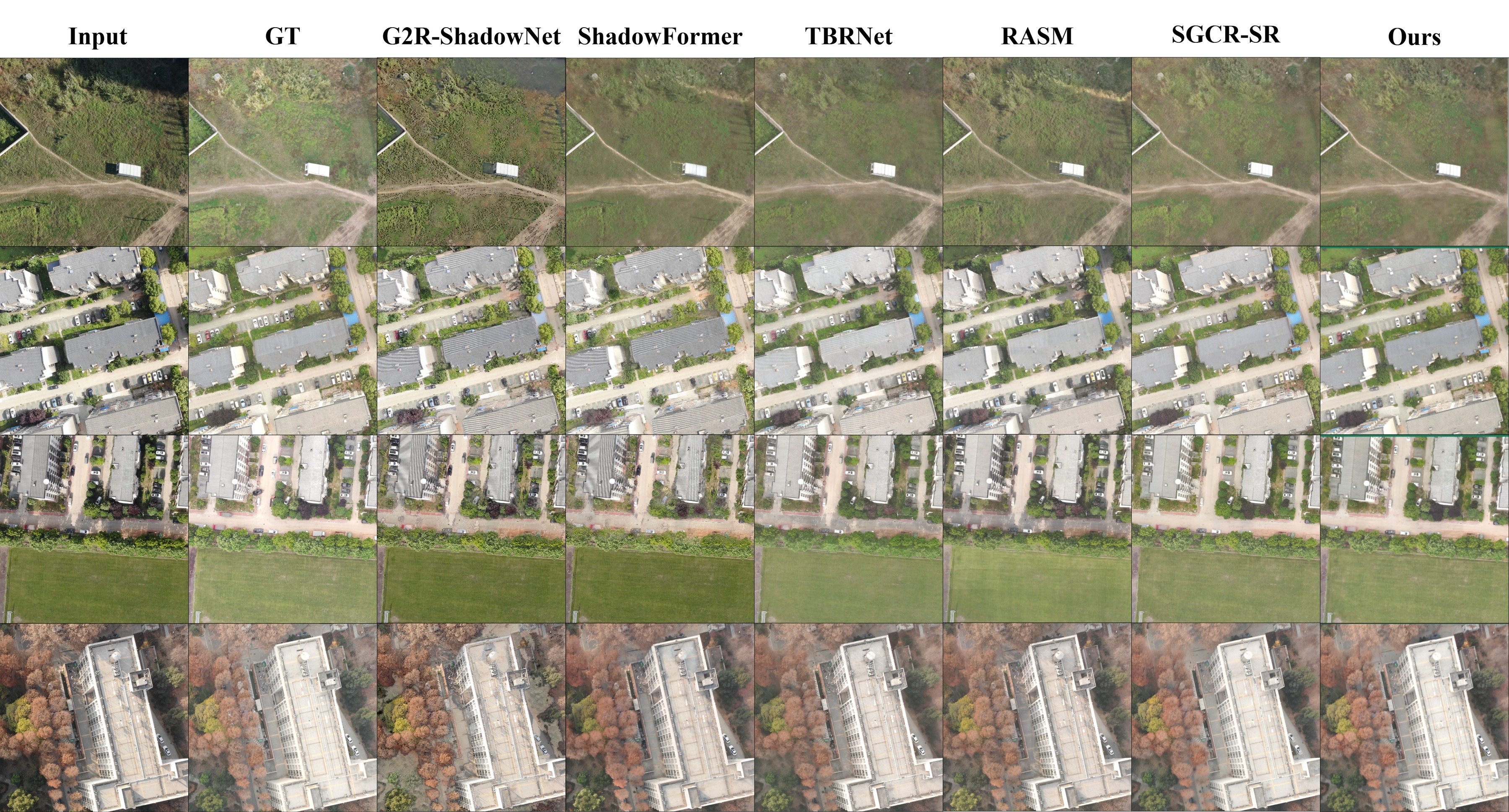}
  \caption{Visual comparison of shadow removal on the UAV-SC+ dataset.}
  \label{fig:qual_uavsc}
\end{figure*}

\begin{figure}[t]
  \centering
  \begin{subfigure}{0.85\linewidth}
    \centering
    \includegraphics[width=\linewidth]{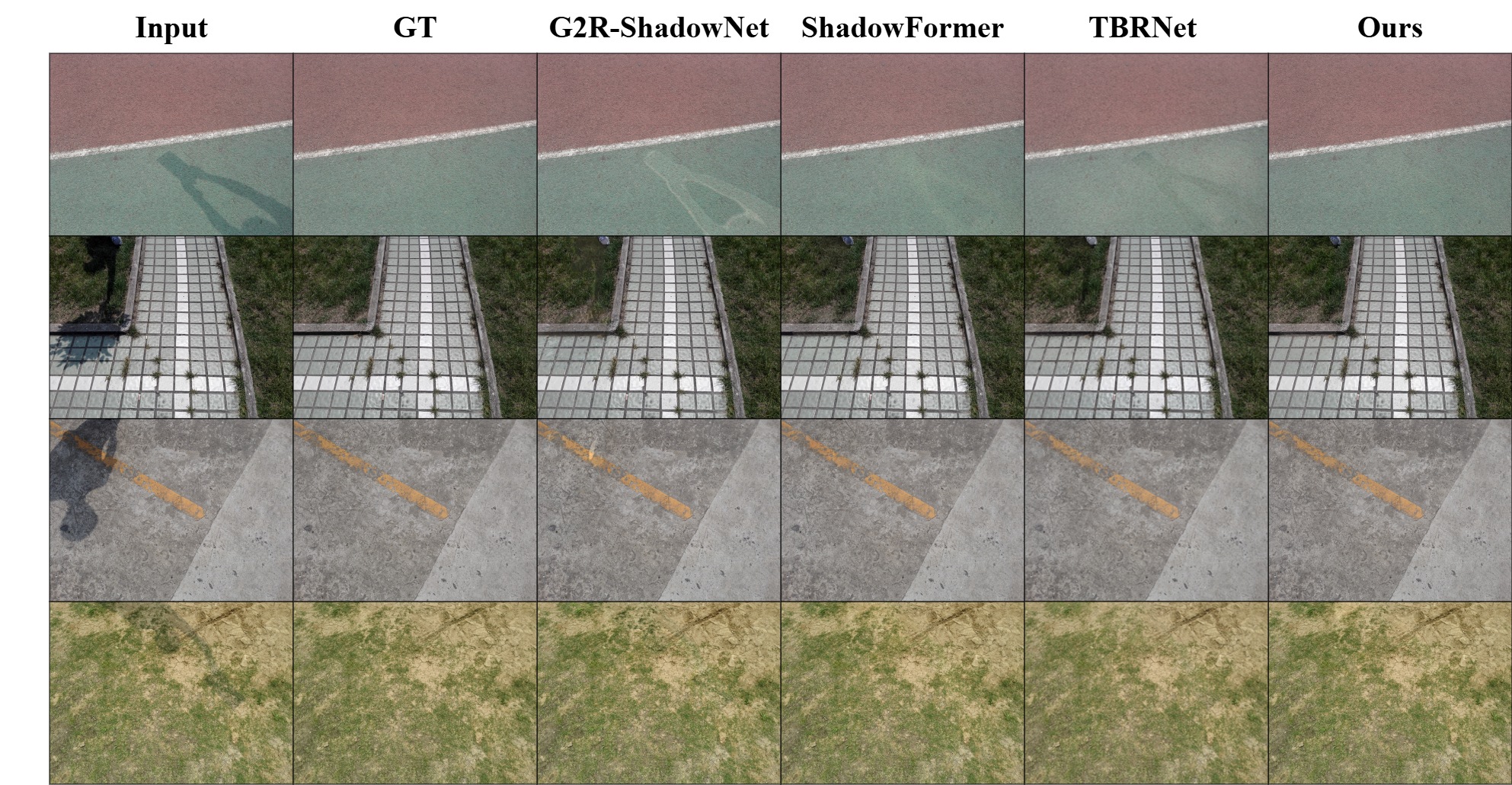}
    \caption{ISTD+ dataset}
  \end{subfigure}
  \vspace{2pt}
  \begin{subfigure}{0.85\linewidth}
    \centering
    \includegraphics[width=\linewidth]{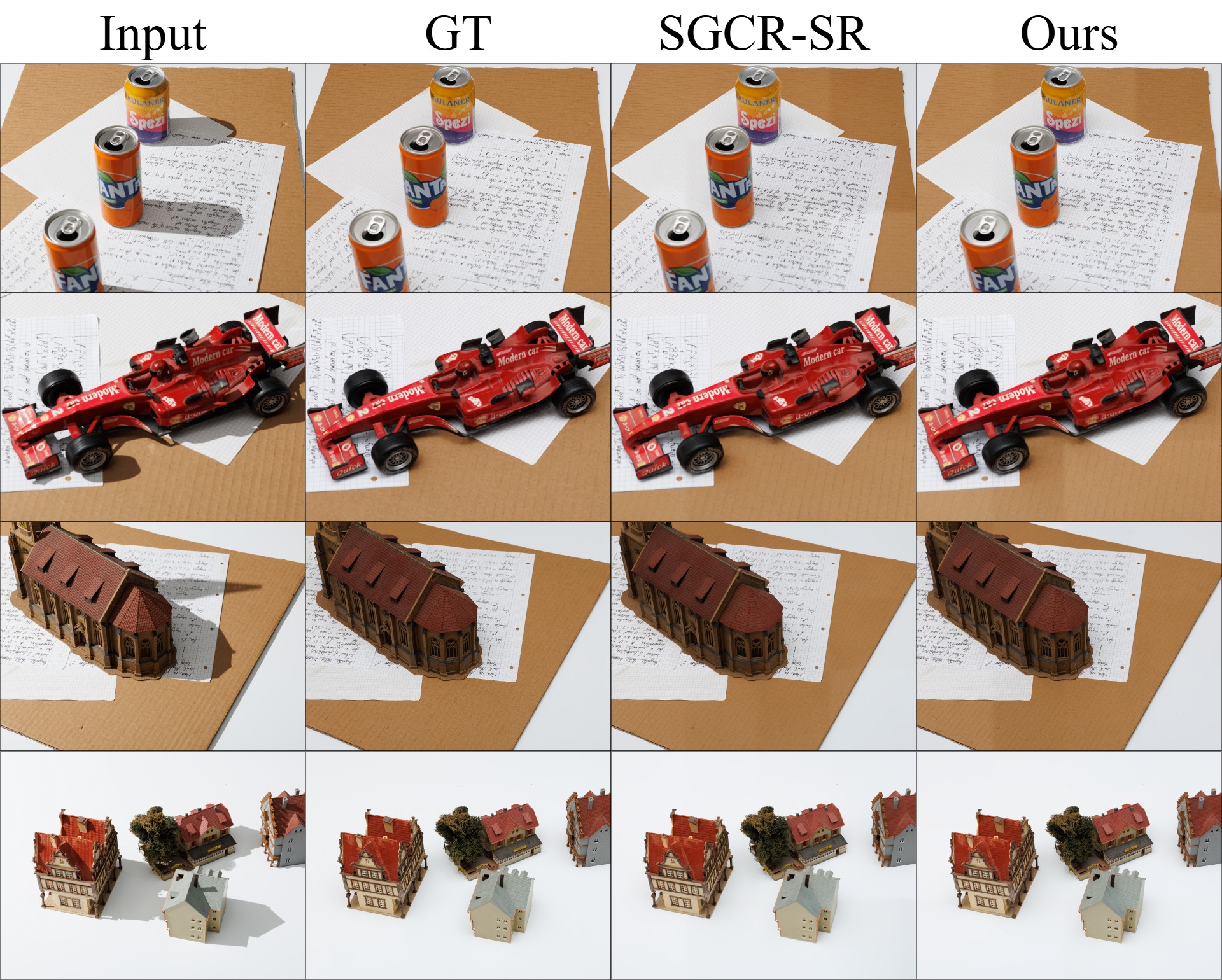}
    \caption{WSRD+ 2026 dataset}
  \end{subfigure}
  \caption{Visual comparison of shadow removal on the ISTD+ and WSRD+ 2026 datasets.}
  \label{fig:qual_istd_wsrd}
\end{figure}

The left part of Table~\ref{tab:uavsc} compares our method with state-of-the-art methods on the UAV-SC+ dataset. We rank first in whole-image PSNR (26.65 dB), and the PSNR in both the shadow and non-shadow regions is substantially improved (34.57 dB and 28.08 dB), which shows that our method not only improves the restoration quality of the shadow region but also effectively alleviates the degradation of the non-shadow region. Meanwhile, SSIM and LPIPS reach 0.897 and 0.0772, respectively, indicating that the mask prior better preserves region boundaries and texture details. The shadow removal results on UAV-SC+ are shown in Fig.~\ref{fig:qual_uavsc}.

The right part of Table~\ref{tab:uavsc} compares our method with 8 representative methods on the ISTD+ dataset. Our method achieves the best results on all five metrics: the shadow-region and non-shadow-region PSNR are improved by 1.31 dB and 1.63 dB over the best respective baselines, and whole-image PSNR, SSIM and LPIPS also lead across the board. The other methods exhibit a clear trade-off between the two types of regions, whereas our method is optimal in both regions simultaneously, showing that the improvement in shadow removal quality achieved by cascaded responsibility separation is not limited to remote sensing images but also holds for natural images; the simultaneous improvement of SSIM and LPIPS also indicates that our method improves both structural fidelity and perceptual quality. Fig.~\ref{fig:qual_istd_wsrd}(a) shows the visual comparison on the ISTD+ dataset.

\begin{center}
  \setlength{\tabcolsep}{1.4pt}
  \captionof{table}{On the WSRD+ 2026 dataset, we compare our method with the baseline SGCR-SR under identical test conditions. PSNR is reported on the shadow region, the non-shadow region and the whole image, respectively.}
  \label{tab:wsrd}
  \vspace{2pt}
  \begin{tabular*}{\linewidth}{@{\extracolsep{\fill}}lccccc@{}}
    \toprule
    \multirow{2}{*}{Method} & \multicolumn{3}{c}{PSNR (dB) $\uparrow$} & \multirow{2}{*}{SSIM $\uparrow$} & \multirow{2}{*}{LPIPS $\downarrow$} \\
    \cmidrule(lr){2-4}
     & S. & N.S. & All & & \\
    \midrule
    SGCR-SR \cite{ref9} & 28.816 & 29.066 & 28.964 & 0.8992 & 0.0769 \\
    Ours                & \textbf{30.618} & \textbf{29.738} & \textbf{29.796} & \textbf{0.9029} & \textbf{0.0420} \\
    \bottomrule
  \end{tabular*}
  \end{center}

Table~\ref{tab:wsrd} compares our method with SGCR-SR, the winning solution of the CVPR 2026 NTIRE Image Shadow Removal Challenge. On the WSRD+ 2026 dataset, our method improves the PSNR on the shadow region, non-shadow region and whole image by 1.80 dB, 0.67 dB and 0.83 dB, respectively, improves SSIM by 0.0037 and reduces LPIPS by 0.0349, outperforming the winning solution on all metrics. Fig.~\ref{fig:qual_istd_wsrd}(b) shows the visual comparison on the WSRD+ 2026 dataset.

\subsection{Ablation Study}
\label{sec:exp:ablation}

We propose three modules: the responsibility separation strategy (STARS), the feature selection and fusion adapter (FSFA) and the non-shadow region-specific loss ($\mathcal{L}_\mathrm{ns}$). To verify the effectiveness of each module, we conduct a module-wise ablation on the UAV-SC+ dataset; the results are shown in Table~\ref{tab:ablation}. To ensure a fairer comparison, all metrics in the table are computed at the whole-image resolution. Therefore, the results of the full model differ noticeably from those in Table~\ref{tab:uavsc}.

\begin{center}
  \setlength{\tabcolsep}{1.4pt}
  \captionof{table}{Ablation results on the UAV-SC+ dataset.}
  \label{tab:ablation}
  \vspace{2pt}
  \begin{tabular*}{\linewidth}{@{\extracolsep{\fill}}lccccc@{}}
    \toprule
    \multirow{2}{*}{Method} & \multicolumn{3}{c}{PSNR (dB) $\uparrow$} & \multirow{2}{*}{SSIM $\uparrow$} & \multirow{2}{*}{LPIPS $\downarrow$} \\
    \cmidrule(lr){2-4}
     & S. & N.S. & All & & \\
    \midrule
    Full model                     & \textbf{33.74} & \textbf{26.58} & \textbf{25.35} & \textbf{0.817} & 0.1743 \\
    w/o STARS                      & 33.70 & 26.48 & 25.26 & 0.806 & \textbf{0.1683} \\
    w/o FSFA                       & 32.68 & 25.30 & 24.01 & 0.804 & 0.1922 \\
    w/o $\mathcal{L}_\mathrm{ns}$  & \textbf{33.74} & 26.55 & 25.32 & 0.805 & 0.1686 \\
    \bottomrule
  \end{tabular*}
  \end{center}

In Table~\ref{tab:ablation}, removing FSFA decreases the three PSNR values by 1.06/1.28/1.34 dB and degrades LPIPS from 0.1743 to 0.1922, which is the most obvious degradation, indicating that the large-model semantic prior is fundamental to cascade restoration; removing STARS decreases the whole-image PSNR by 0.09 dB and SSIM by 0.011, and the loss in the non-shadow region (0.10 dB) is larger than that in the shadow region (0.04 dB), consistent with its role of suppressing non-shadow degradation, while removing $\mathcal{L}_\mathrm{ns}$ shows a similar but smaller trend, indicating that the structural division indeed requires the accompanying optimization signal to be fully executed. The full model is optimal in PSNR and SSIM, while its LPIPS is slightly higher than that of the variants without STARS or $\mathcal{L}_\mathrm{ns}$; this is because LPIPS is more sensitive to high-frequency texture, whereas the two modules suppress over-correction in the non-shadow region and make the output smoother. Its difference in LPIPS is far smaller than the fidelity gain, corresponding to a slight trade-off between fidelity and perceptual metrics.

\begin{figure}[t]
  \centering
  \includegraphics[width=\linewidth]{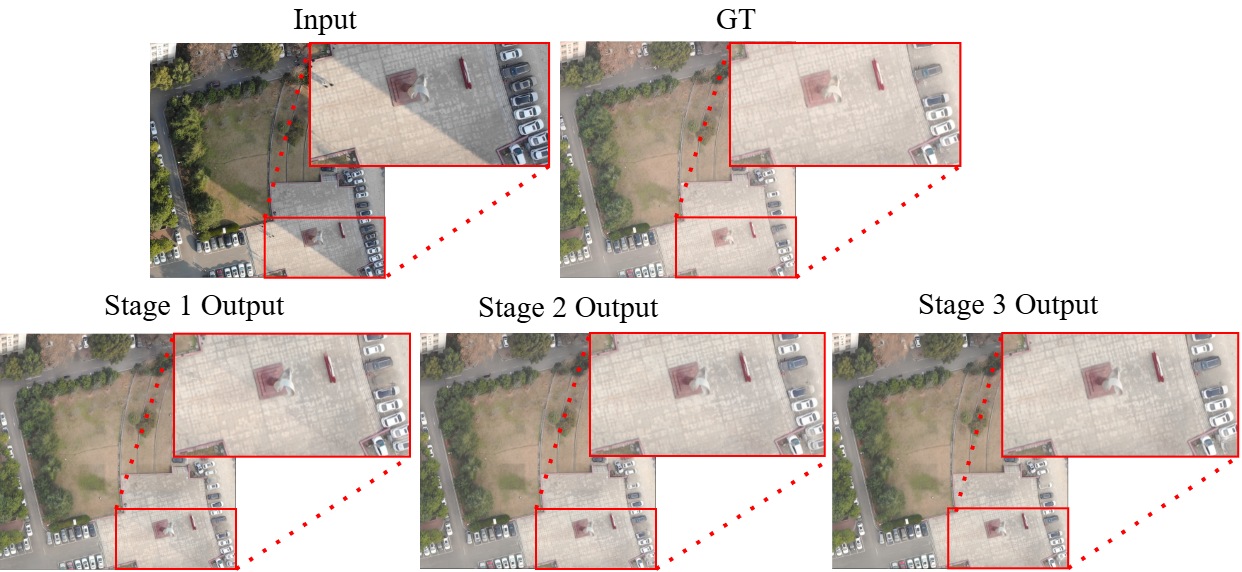}
  \caption{Stage-wise restoration results.}
  \label{fig:stagewise}
\end{figure}

\paragraph{Stage-wise Restoration Results}
We extract the outputs of the individual stages inside the cascade and measure their PSNR, SSIM and LPIPS, so as to compare the restoration capability of the stages vertically. The results are listed in Table~\ref{tab:stagewise}, and Fig.~\ref{fig:stagewise} provides the visual illustration.

\begin{center}
  \setlength{\tabcolsep}{1.4pt}
  \captionof{table}{Outputs of each stage extracted from the full model trained on the UAV-SC+ dataset, compared with each other.}
  \label{tab:stagewise}
  \vspace{2pt}
\begin{tabular*}{\linewidth}{@{\extracolsep{\fill}}lccccc@{}}
    \toprule
    \multirow{2}{*}{Stage} & \multicolumn{3}{c}{PSNR (dB) $\uparrow$} & \multirow{2}{*}{SSIM $\uparrow$} & \multirow{2}{*}{LPIPS $\downarrow$} \\
    \cmidrule(lr){2-4}
     & S. & N.S. & All & & \\
    \midrule
    Stage 1 & 33.79 & 27.13 & 25.74 & 0.884 & 0.0855 \\
    Stage 2 & 34.37 & 27.78 & 26.35 & 0.893 & 0.0800 \\
    Stage 3 & 34.58 & 28.04 & 26.62 & 0.897 & 0.0772 \\
    \bottomrule
  \end{tabular*}
  \end{center}

As can be seen from Table~\ref{tab:stagewise}, as the cascade proceeds, the PSNR in the shadow region, non-shadow region and whole image increases monotonically, SSIM rises from 0.884 to 0.897, and LPIPS decreases from 0.0855 to 0.0772, indicating that each stage brings a positive gain to the final result. The improvement from Stage 1 to Stage 2 is the largest, suggesting that the first two stages mainly handle the restoration of the main content; the gain from Stage 2 to Stage 3 is relatively smaller, but the non-shadow region still improves consistently, which is consistent with the role of the last stage in correcting the non-shadow region. These stage-wise gains show that the cascade is not merely repeated refinement, but lets different stages take different regional objectives and accumulate the restoration effect progressively.

\section{Discussion}

We propose STARS, a responsibility separation strategy built on a cascaded architecture, which separates shadow removal for shadow and non-shadow regions into different stages, and we design a feature selection and fusion adapter (FSFA) that introduces frozen large-model semantic priors in a lightweight manner. On UAV-SC+, ISTD+ and WSRD+ 2026, our method achieves state-of-the-art performance; quantitative and qualitative evaluations show that explicitly dividing regional responsibilities is an effective way to alleviate the side effects of mask priors and improve both regions simultaneously.

Our method is based on a three-stage cascade and still requires higher time and computational cost; in future work, we will explore more efficient designs, such as reusing inter-stage parameters or adopting lighter semantic guidance, to improve practicality.

\section*{LLM Usage Statement}

The authors used large language models (LLMs), DeepSeek and ChatGPT, during the preparation of this paper. Specifically, the LLMs were used for: (i) language polishing of the paragraphs that analyze the experimental results; (ii) translating the manuscript from Chinese into English; (iii) LaTeX typesetting and formatting of the paper; (iv) generating figures from the authors' experimental data; and (v) collecting the baseline numbers of prior methods from their published papers for the comparison tables, such as the ISTD+ results in Table~1. The LLMs did not contribute research ideas, the proposed method, or the experimental results, and the assisted content is derived from the authors' own work. The authors have checked all LLM-assisted text, numbers, figures and citations against the original sources, and take full responsibility for the content of this paper.

\end{document}